\pdfoutput=1

\documentclass[11pt]{article}

\usepackage{ACL2023}

\usepackage{times}
\usepackage{latexsym}
\usepackage{graphicx}
\usepackage{subcaption}
\usepackage{makecell}
\usepackage{hhline}

\usepackage[T1]{fontenc}

\usepackage[utf8]{inputenc}

\usepackage{microtype}

\usepackage{inconsolata}

\usepackage{adjustbox} 
\usepackage{booktabs}
\usepackage{multirow}
\usepackage{colortbl}

\title{Guidance-Based Prompt Data Augmentation in Specialized Domains for Named Entity Recognition}

\author{
Hyeonseok Kang$^1$, Hyein Seo$^1$, Jeesu Jung$^1$, Sangkeun Jung$^1$\thanks{\ \ Corresponding author}, Du-Seong Chang$^2$, Riwoo Chung$^2$ \\
$^1$Computer Science and Engineering, Chungnam National University, Republic of Korea \\
$^2$KT Corporation, Republic of Korea\\
\texttt{\{dnfldjaak11,hyenee97,jisu.jung5,hugmanskj\}@gmail.com},
\texttt{\{dschang,riwoo.chung\}@kt.com},
}

\begin{document}
\maketitle
\begin{abstract}
While the abundance of rich and vast datasets across numerous fields has facilitated the advancement of natural language processing, sectors in need of specialized data types continue to struggle with the challenge of finding quality data. Our study introduces a novel \textit{guidance data augmentation} technique utilizing abstracted context and sentence structures to produce varied sentences while maintaining context-entity relationships, addressing data scarcity challenges. By fostering a closer relationship between context, sentence structure, and role of entities, our method enhances data augmentation's effectiveness. Consequently, by showcasing diversification in both entity-related vocabulary and overall sentence structure, and simultaneously improving the training performance of named entity recognition task.
\end{abstract}

\section{Introduction}

The field of Natural Language Processing (NLP) has witnessed remarkable success across various domains in recent years, primarily attributed to the availability of rich and high-quality data. However, specialized fields such as science and biology face significant challenges due to the scarcity of such quality data. Particularly, tasks like Named Entity Recognition (NER) face significant difficulties due to domain-specific characteristics where vocabulary roles diverge from general usage, necessitating specialized knowledge for effective data collection.
To overcome the data shortage issue, various automated data augmentation (DA) techniques have been developed, including a recent approach that leverages Large Language Models (LLMs) for sentence generation to perform DA \cite{llmDA}.
Utilizing LLMs for DA involves employing few-shot learning or external modules \cite{toolqa} to provide additional information. In NER tasks, DA is applied with a focus on entities, maintaining the sentence's core structure with minimal alterations. This approach faces limitations in effectively augmenting cases like specific domain data, where vocabulary interpretation and roles vary with context and sentence composition.

In this study, we propose \textbf{G}uidance \textbf{D}ata \textbf{A}ugmentation (\textbf{GDA}), utilizing information on context and sentence structure abstracted through data abstraction for DA, aiming to generate sentences with varied structures alongside augmenting similar entity types. This approach seeks to achieve more natural and diverse DA compared to single LLM methods by augmenting data with sentences of varied structures that match the seed sentence's context and corresponding entity types.

\begin{figure}[t!]
    \centering
    \includegraphics[width=0.48\textwidth]{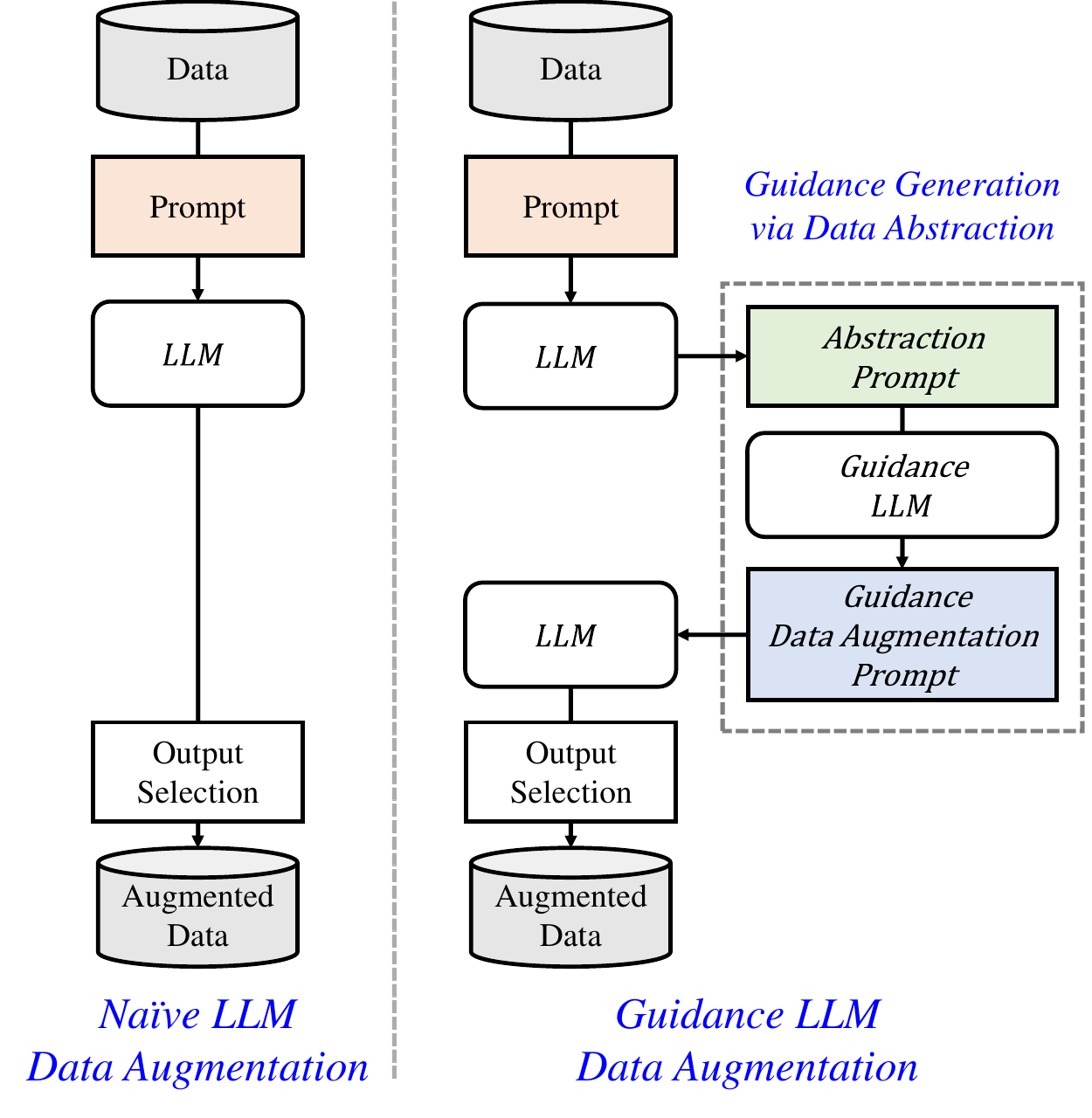}
    \caption{Comparison between data augmentation using LLM and Guidance LLM-based data augmentation.}
    \label{fig:framework}
\end{figure}

\begin{figure*}[t!]
    \centering
    \includegraphics[width=1.0\textwidth]{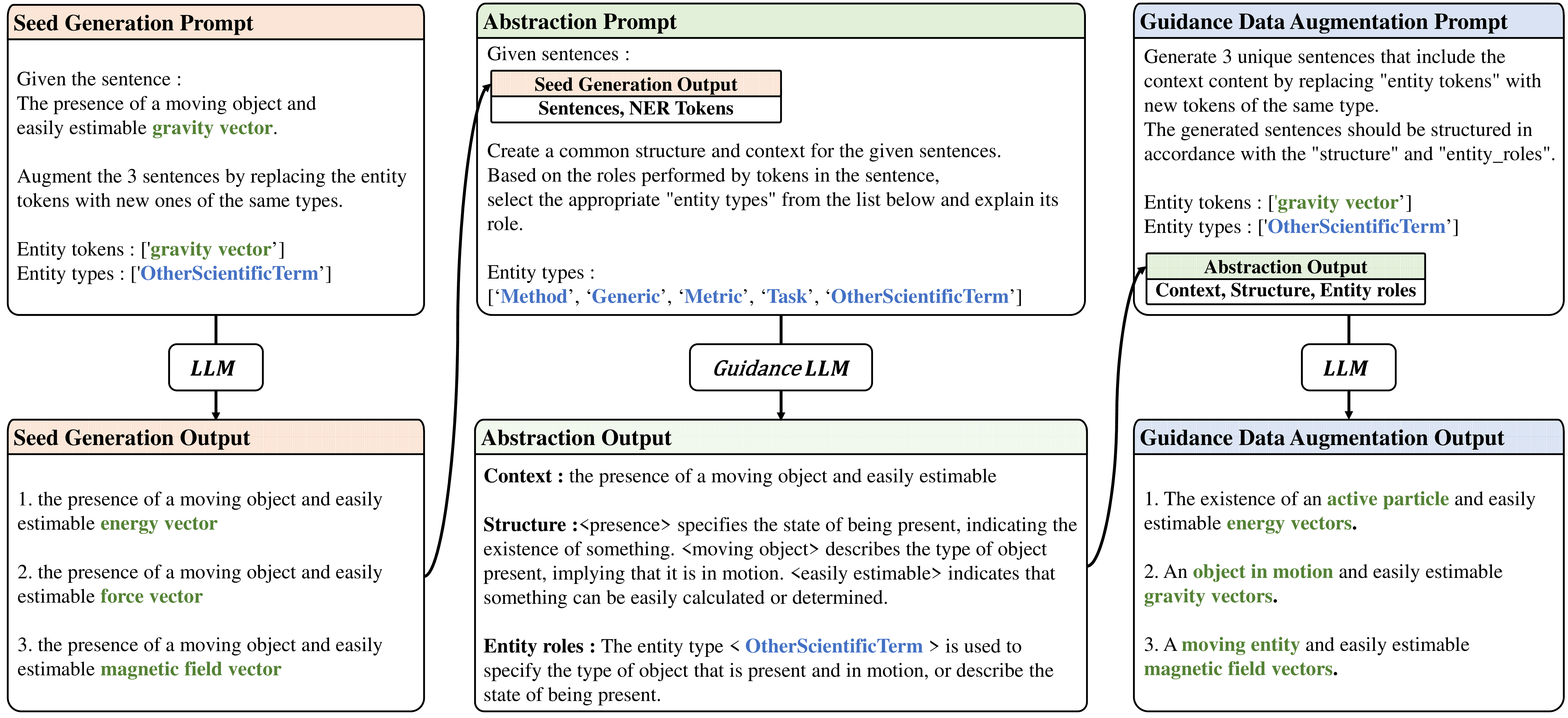}
    \caption{Illustration of prompt flows for data augmentation in NER tasks. Structuring the use of data augmentation prompts with guidance prompts. Unlike traditional methods that input only the named entity tokens and their types, guidance prompts utilize context and entity role information generated from abstraction prompts to enrich data augmentation.}
    \label{fig:prompt_overview}
\end{figure*}

Our data abstraction approach structures relationships among context, entities, and sentence composition for DA, expanding inference scope by using higher-level conceptual information\cite{StepBack}. This approach is vital where entity-related terms diverge significantly from general usage, requiring a sophisticated strategy that analyzes vocabulary and structural details of entities. We address this challenge with our data abstraction approach, which assesses the contextual environment, the relationships between entities and their context, and the roles entities play, ensuring the development of varied and contextually appropriate terminology for specialized areas.

In response to these considerations, our study introduces a guidance prompt-based DA framework (see Figure \ref{fig:framework}). This framework is designed to generate sentences of various structures using the same type of entity, from data abstraction to the final response selection.


\section{Related Works}
In NLP, widely used DA techniques comprise rule-based approaches such as synonym replacement, back-translation, and random text element insertion or deletion. These methods are especially prevalent in tasks where textual data may require diversification to better train models \cite{Bayer:DA}.  Specifically, for tasks such as NER, augmentation strategies often revolve around the substitution of words with similar meanings or roles. In this context, techniques like Easy Data Augmentation (EDA) \cite{EDA} and the utilization of WordNet \cite{wordnet} through the Natural Language Toolkit (NLTK) \cite{nltk} are frequently applied to generate synonyms-based augmented data.
In particular, when using data from specialized domain, DA methods are often used because it is difficult to collect data as it often consists of data containing domain-specific knowledge. For biomedical named entity recognition, augmentation is often performed using context to enhance the understanding of specialized concepts \cite{bioDA}.
Recently, the utilization of LLMs has expanded, leading to an increased use of DA techniques based on LLMs. These techniques involve augmenting data for sentence classification by leveraging LLMs \cite{chataug}, or enhancing cross-lingual tasks \cite{crosslingual} through augmentation.

\section{Guidance Data Augmentation} \label{method}
Our framework is designed around two key components for effective DA aimed at NER tasks: \textit{\textbf{Data Abstraction}} and \textit{\textbf{Data Augmentation via Guidance Prompts}}. Through these two approaches, the proposed method facilitates enhanced model performance on NER tasks.

\subsection{Guidance as Data Abstraction}
Data abstraction involves abstracting and generalizing data to a form where the essential qualities are retained without the unnecessary specifics. The process allows for the alignment of roles and contextual attributes of named entities within sentences with the required entity types for the NER task, thereby enabling the systematic identification and extraction of pivotal information.

Initially, for the purpose of data abstraction, a prompt is constructed to process data by substituting tokens corresponding to named entities with alternative tokens of the same type, as depicted in the seed generation prompt in Figure \ref{fig:prompt_overview}. The data generated from these prompts, along with seed data and a comprehensive list of entity types, are used to formulate abstraction prompts. Such abstracted prompts are fed into a guidance LLM to generate common contextual information for sentences and to produce the necessary structure and entity role information for context composition.

\subsection{Data Augmentation via Guidance Prompt}
Following data abstraction, the second component focuses on the augmentation process itself, using guidance prompts to generate new and varied instances of text. This method utilizes the abstracted data as a basis to inform the generation process, ensuring that the newly created text is both relevant and diverse. The guidance prompts are designed to direct the LLM in producing sentences that not only contain the targeted named entities but also mirror the semantic and structural diversity found in natural language usage.

The final output generated using guidance prompts is configured to create sentences that include the semantic information of the context without the need for seed data. The entities' roles and structure, derived from data abstraction alongside the context, serve as essential information for the generation and assignment of named entities. These elements are incorporated as conditions that must be adhered to within the guidance prompt, ensuring that the generated text aligns with specified contextual meanings and structural requirements.

\begin{table}
\begin{centering}
\begin{adjustbox}{max width=\linewidth}
\begin{tabular}{cccccc}
\toprule 
~  & \textbf{SciERC} & \textbf{NCBI-disease} & \textbf{FIN} \\
\hline 
Train & 1,861\small(200) & 5,432\small(200) & 1,018\small(200) \tabularnewline
Dev & 275 & 923 & 150 \tabularnewline
Test & 551 & 940 & 305 \tabularnewline
\bottomrule 
\end{tabular}
\end{adjustbox}
\par\end{centering}
\caption{\label{tab:data_statistics}Composition of the dataset for model evaluation. Values within parentheses in the train dataset column represent the number of seed data instances randomly selected from the training data for augmentation purposes.}
\end{table}

Together, these two components offer an effective approach to enhancing the training data for NER tasks. By first abstracting the data to capture its essential elements and then augmenting it through carefully designed prompts, our framework aims to significantly diversify the datasets available for training NER models.


\section{Experiments}
\subsection{Datasets}
In selecting datasets for our experiments, we focused on data from specialized domains. These domains are characterized by the specificity (or expertise) required in their entities, requiring specialized knowledge for DA. Specifically, only three datasets were used: SciERC, NCBI-disease, and FIN. Detailed descriptions of the datasets are provided in Appendix \ref{sec:dataset_detail}. 




\begin{table*}
\centering
\begin{adjustbox}{max width=\linewidth}
\begin{tabular}{llcccc}
\hline
\multirow{2}{*}{\textbf{Approch}} & \multirow{2}{*}{\textbf{Method}} & \multirow{2}{*}{\textbf{Model}} & \multicolumn{3}{c}{\textbf{Datasets}} \\
\cline{4-6}
~ & ~ & ~ & \textbf{SciERC} & \textbf{NCBI-disease} & \textbf{FIN} \\
\hline
\multirow{2}{*}{\makecell{Rule-based\\DA}} & WordNet     & \multirow{2}{*}{-} & 0.5018 & 0.7924 & 0.7480   \\
~ & EDA         & ~ & 0.5434 & 0.8062 & 0.7953  \\
\hline
\multirow{4}{*}{\makecell{LLM-based\\DA}} & Naïve DA   & \multirow{2}{*}{GPT-3.5} & 0.5342(\small\textcolor{red}{-0.0092}) & 0.8017(\small\textcolor{red}{-0.0045}) & 0.8440(\small\textcolor{blue}{+0.0480)} \\
~ & GDA(Ours)         &   ~ ~   & \cellcolor{cyan!20}{\textbf{0.5435}(\small\textcolor{blue}{+0.0001})} & \cellcolor{cyan!20}{\textbf{0.8139}(\small\textcolor{blue}{+0.0077})} & \cellcolor{cyan!20}{0.8464(\small\textcolor{blue}{+0.0511})} \\
\hhline{~|*5-}
~ & Naïve DA   & \multirow{2}{*}{GPT-4} & \cellcolor{cyan!20}{0.5308(\small\textcolor{red}{-0.0126})} & 0.7697(\small\textcolor{red}{-0.0365}) & 0.8520(\small\textcolor{blue}{+0.0567}) \\
~ & GDA(Ours)         &   ~ ~ & 0.5159(\small\textcolor{red}{-0.0275}) & \cellcolor{cyan!20}{0.7875(\small\textcolor{red}{-0.0187})} & \cellcolor{cyan!20}{\textbf{0.8544}(\small\textcolor{blue}{+0.0591})}  \\
\hline
\end{tabular}
\end{adjustbox}
\caption{\label{tab:ner_evaluation}
Evaluation of models trained with augmented data. Utilizing a base of 200 seed data points to generate an additional 600 data points for training, resulting in a total dataset of 800 entries. The table highlights the augmentation technique yielding the highest F1 score for each dataset in \textbf{bold}. Baseline methods employed were WordNet and EDA, with scores in parentheses indicating F1 score comparisons based on EDA. For LLM-based data augmentation, methods with superior performance per model are highlighted with a \colorbox{cyan!20}{cyan} background.}
\end{table*}

\subsection{Models}
\textbf{Data augmentation LLMs} Within the proposed framework, the following LLMs were used OpenAI GPT-3.5 and GPT-4 \cite{RLHF}. Models versioned gpt-3.5-turbo-0125 and gpt-4-0613 were utilized, with the temperature parameter set to the default value of 1. By using the same LLM version for both abstraction and augmentation output generation in guidance data augmentation, this approach prevents the influence of language understanding differences among LLMs. Detailed information on the DA setting employing LLM is delineated in Appendix \ref{sec:baseline}.


\noindent 
\textbf{Evaluation model for NER task} For the evaluation phase, we used pre-trained language models, specifically BERT \cite{bert2019kenton}. The model version employed is bert-base-uncased, and a comparative study was conducted to analyze the training effects of DA methods across three datasets. For training, the model was fed with a combination of 200 seed data instances and data augmented through DA as the training dataset. The F1 score was utilized as the metric for evaluating NER task performance. The implementation details utilized for training the evaluation model are furnished in Appendix \ref{sec:imple_detail}.


\subsection{Results}
Table \ref{tab:ner_evaluation} presents a comparison of NER model performance, contrasting models fine-tuned with datasets augmented using baseline methods such as EDA, WordNet, and Naïve DA with those augmented through the proposed GDA approach. When employing GPT-3.5 for data augmentation, the proposed method generally outperformed the Naïve DA approach. Notably, the NCBI-disease dataset saw a 1.22\% improvement in F1 score with the proposed method over Naïve DA, while the FIN dataset experienced a 0.24\% increase. Augmentation with GPT-4 yielded a 1.78\% and 0.24\% performance boost for the NCBI-disease and FIN datasets, respectively.


\begin{figure}[t!]
    \centering
    \begin{subfigure}{0.48\textwidth}
        \centering
        \includegraphics[width=1.0\textwidth]{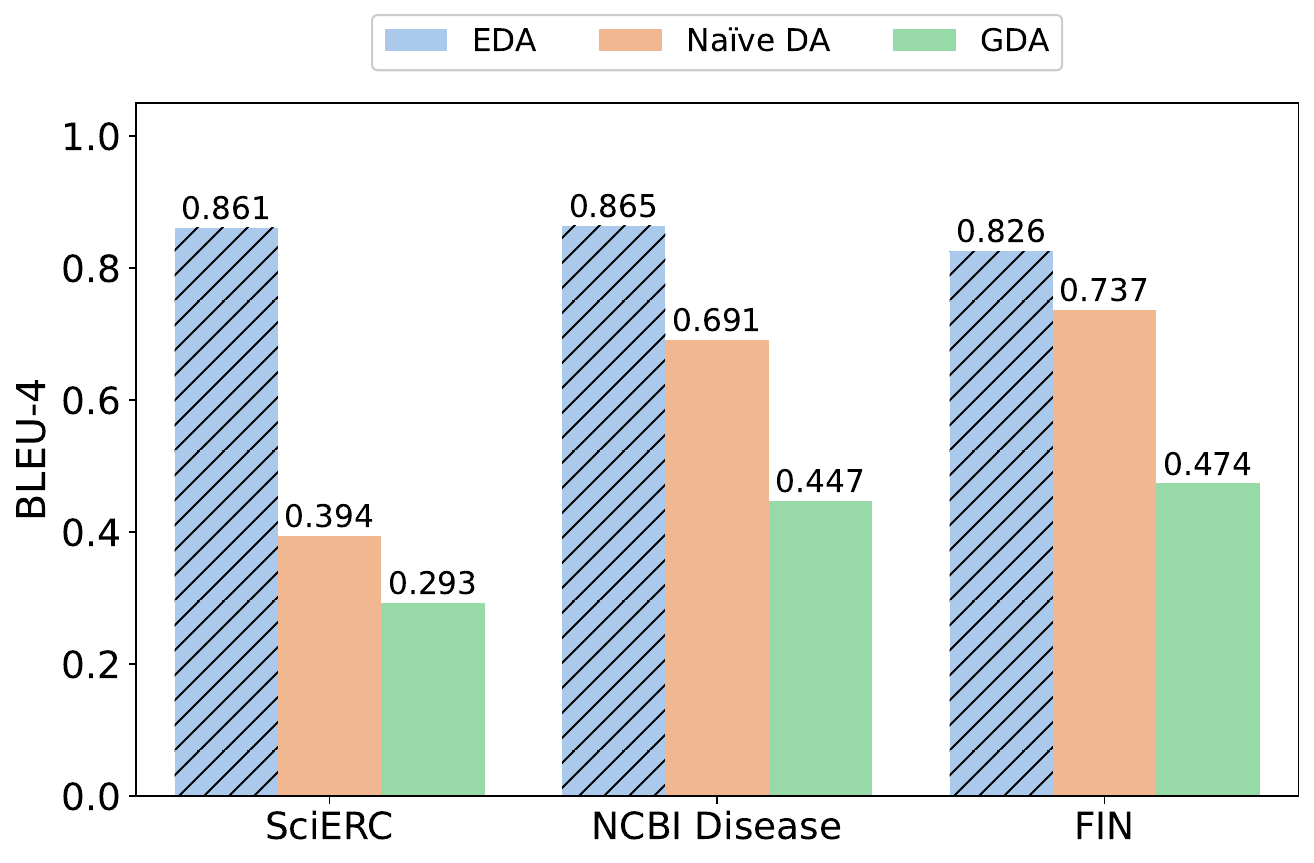}
        \caption{BLEU-4 in augmentation using GPT-3.5}
        \label{fig:bleu_scoring_gpt3.5}
    \end{subfigure}
    \hfill 
    \begin{subfigure}{0.48\textwidth}
        \centering
        \includegraphics[width=1.0\textwidth]{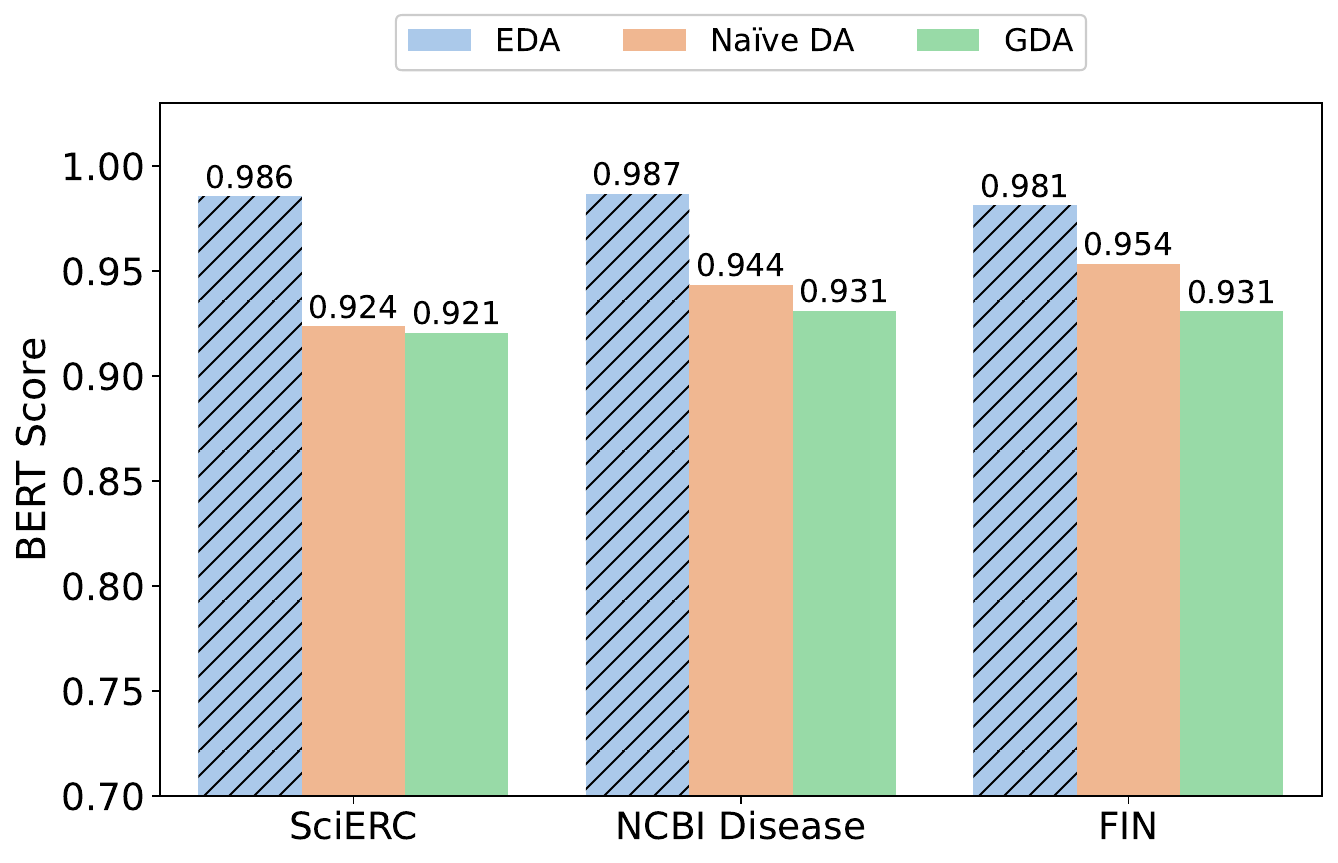}
        \caption{BERTScore in augmentation using GPT-3.5}
        \label{fig:bert_scoring_gpt3.5}
    \end{subfigure}
    \caption{Comparison of sentence diversity by method in augmentation using GPT-3.5 model. Lower BLEU and BERTScore indicate higher diversity in generated sentences.}
    \label{fig:scoring_comparison_gpt3.5}
\end{figure}

In addition to improving model training performance, enhancing data augmentation requires the generation of diverse sentence structures and the assembly of vocabulary that corresponds with entity types. Figure \ref{fig:bleu_scoring_gpt3.5} displays a graph comparing the structure of sentences generated through DA by method, utilizing the Bilingual Evaluation Understudy (BLEU) \cite{bleu} score to assess the degree of n-gram match between two sentences. Evaluation was conducted using BLEU-4, where a lower score signifies reduced structural similarity between sentences, indicating greater diversity in the generated sentences. 

In addition to BLEU-4, we utilized BERTScore \cite{bertscore} to evaluate the semantic similarity of generated sentences. BERTScore is particularly useful for capturing the nuanced semantic differences between sentences, as it utilizes pre-trained transformer models to provide a more context-aware assessment of similarity. This method is advantageous over traditional n-gram based metrics because it can better account for the semantic context rather than just surface-level text similarity.
All three datasets displayed lower scores in both BERTScore and BLEU when using LLM-based augmentation compared to EDA, indicating enhanced diversity. In BLEU scores, the SciERC dataset, in particular, showed a significant 56.8\% lower score with the proposed method compared to EDA. When contrasted with Naïve DA, the FIN and NCBI-disease datasets recorded 26.3\% and 24.4\% lower scores, respectively. For BERTScore, although the differences were smaller, ranging from 6.5\% to 5\%, the scores were still lower compared to EDA, indicating that the generated sentences maintained semantic diversity. The narrow score differences in semantic similarity comparisons indicate that the generated sentences successfully preserve the contextual meaning of the seed sentences while introducing diversity.
Experiments involving structural and semantic similarity using GPT-4, as well as assessments in generated sentences, are documented in Appendix \ref{sec:gpt4_bleu}. Examples of DA outputs are provided in Appendix \ref{sec:DA_case}. These results underscore the capability of the augmentation method utilizing abstract information to generate sentences with varied structures while preserving context and entity information.


\section{Conclusion}
In this study, we proposed guidance data augmentation designed for NER tasks within specific domain data, enabling the generation of data suited for these tasks. By using data abstraction, our method facilitates structured relationships among context, entities, and sentence composition, allowing for the generation of sentences with diverse structures while ensuring entity consistency. The abstracted sentence information is utilized in constructing guidance prompts, enabling DA with a rich diversity in vocabulary and sentence structures. Future efforts will aim at refining this process for applicability to additional tasks and exploring the use of multiple guidance LLMs to enrich the abstraction information, thereby enhancing the guidance provided.


\section*{Limitations}
Our study's scope was notably confined to the NER task, limiting the versatility of our guidance prompts and data abstraction processes. This narrow focus restricts our exploration of the framework's potential across various tasks. 
The limitation of employing a singular model approach for both data abstraction and guidance DA restricts the diversity of linguistic insights within our system. Future efforts will aim to broaden the application of our framework by utilizing different LLMs and expanding the range and granularity of data abstraction, thus addressing these limitations and fully leveraging the capabilities of Multi-LLMs structures for enhanced language understanding and generation tasks.


\section*{Acknowledgements}
This work was supported by research fund of Chungnam National University.

\bibliography{anthology,ref}
\bibliographystyle{acl_natbib}

\appendix

\begin{table*}
\centering
\begin{adjustbox}{max width=\linewidth}
\begin{tabular}{cp{8cm}p{4cm}}
\hline
\textbf{Method} & \multicolumn{1}{c}{\textbf{Generated sentence}} & \multicolumn{1}{c}{\textbf{Named entity tokens}}\\
\hline
\multirow{3}{*}{Seed data} & The interlingual approach to MT has been repeatedly advocated by researchers originally interested in natural language understanding who take machine translation to be one possible application.
 & `interlingual approach', `MT', `natural language understanding', `machine translation' \\
\hline
\multirow{9}{*}{EDA} & The interlingual approach to MT has been repeatedly advocated by researchers originally interested in natural language understanding who take machine translation to be one possible applications programme. & `interlingual approach', `MT', `natural language understanding', `machine translation'\\
\cline{2-3}
~ & The interlingual approach to MT has been repeatedly advocated by researchers originally interested in instinctive language understanding who take machine translation to be one possible application. & `interlingual approach', `MT', `natural language understanding', `machine translation'\\
\cline{2-3}
~ & The interlingual approach to MT has been repeatedly advocated by researchers originally interested in natural language understanding who take machine translation to be one potential application. & `interlingual approach', `MT', `natural language understanding', `machine translation'\\
\hline
\multirow{9}{*}{Naïve DA} & The contrastive approach to AI has been repeatedly advocated by researchers originally interested in computer vision who take natural language processing to be one possible application. & `contrastive approach', `AI', `computer vision', `natural language processing'
\\
\cline{2-3}
~ & The symbolic approach to ML has been repeatedly advocated by researchers originally interested in pattern recognition who take deep learning to be one possible application. & `symbolic approach', `ML', `pattern recognition', `deep learning'
\\
\cline{2-3}
~ & The neural approach to NLP has been repeatedly advocated by researchers originally interested in information retrieval who take sentiment analysis to be one possible application. & `neural approach', `NLP', `information retrieval', `sentiment analysis'
\\
\hline
\multirow{9}{*}{GDA(Ours)} & The symbolic, contrastive, neural, statistical, and rule-based approaches to AI, ML, NLP, CV, and NLU have been repeatedly advocated by researchers originally interested in various tasks and applications. & `symbolic, contrastive, neural, statistical, and rule-based approaches', `AI, ML, NLP, CV, NLU'\\
\cline{2-3}
~ & The interlingual approach to MT has been repeatedly advocated by researchers originally interested in natural language understanding which is a possible application.& `interlingual approach', `MT', `natural language understanding' \\
\cline{2-3}
~ & Researchers who were originally focused on natural language understanding have repeatedly advocated the interlingual approach to MT which is one of the potential applications. & `natural language understanding', `interlingual approach', `MT'\\

\hline
\end{tabular}
\end{adjustbox}
\caption{\label{tab:da_case}
Examples of data augmentation outcomes from EDA, Naïve DA, and GDA methodologies.}
\end{table*}

\section{Datasets Details}\label{sec:dataset_detail}
SciERC \cite{SciERC} is a collection of scientific abstract annotated with scientific entities, their relations, and coreference clusters. 
NCBI-disease \cite{nbci_disease} consists of PubMed abstracts fully annotated at the mention and concept level to serve as a research resource for the biomedical natural language processing community.
FIN \cite{fin} is composed of financial agreements publicly disclosed through U.S. Securities and Exchange Commission (SEC) filings.


\begin{figure}[t!]
    \centering
    \begin{subfigure}{0.48\textwidth}
        \centering
        \includegraphics[width=1.0\textwidth]{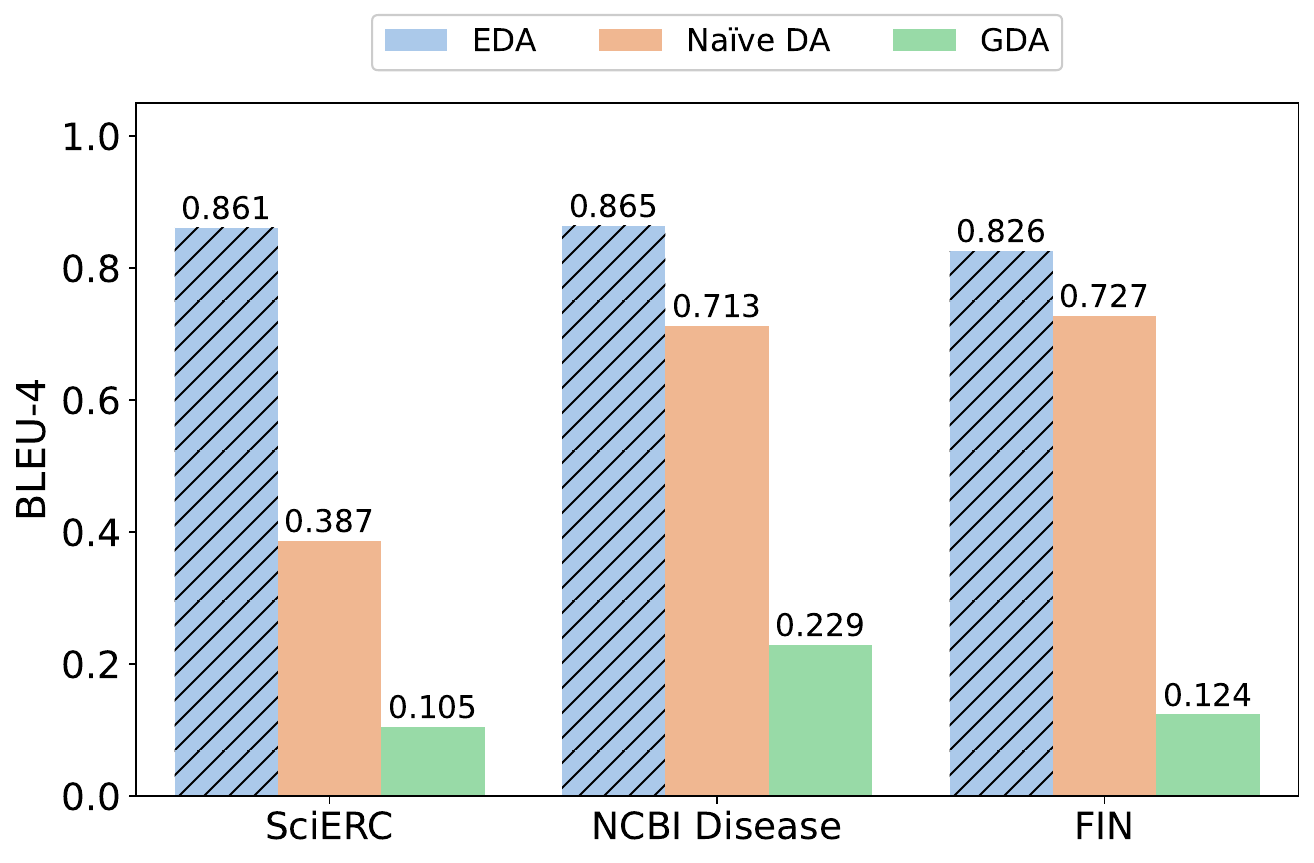}
        \caption{BLEU-4 in augmentation using GPT-4}
        \label{fig:bleu_scoring_gpt4}
    \end{subfigure}
    \hfill 
    \begin{subfigure}{0.48\textwidth}
        \centering
        \includegraphics[width=1.0\textwidth]{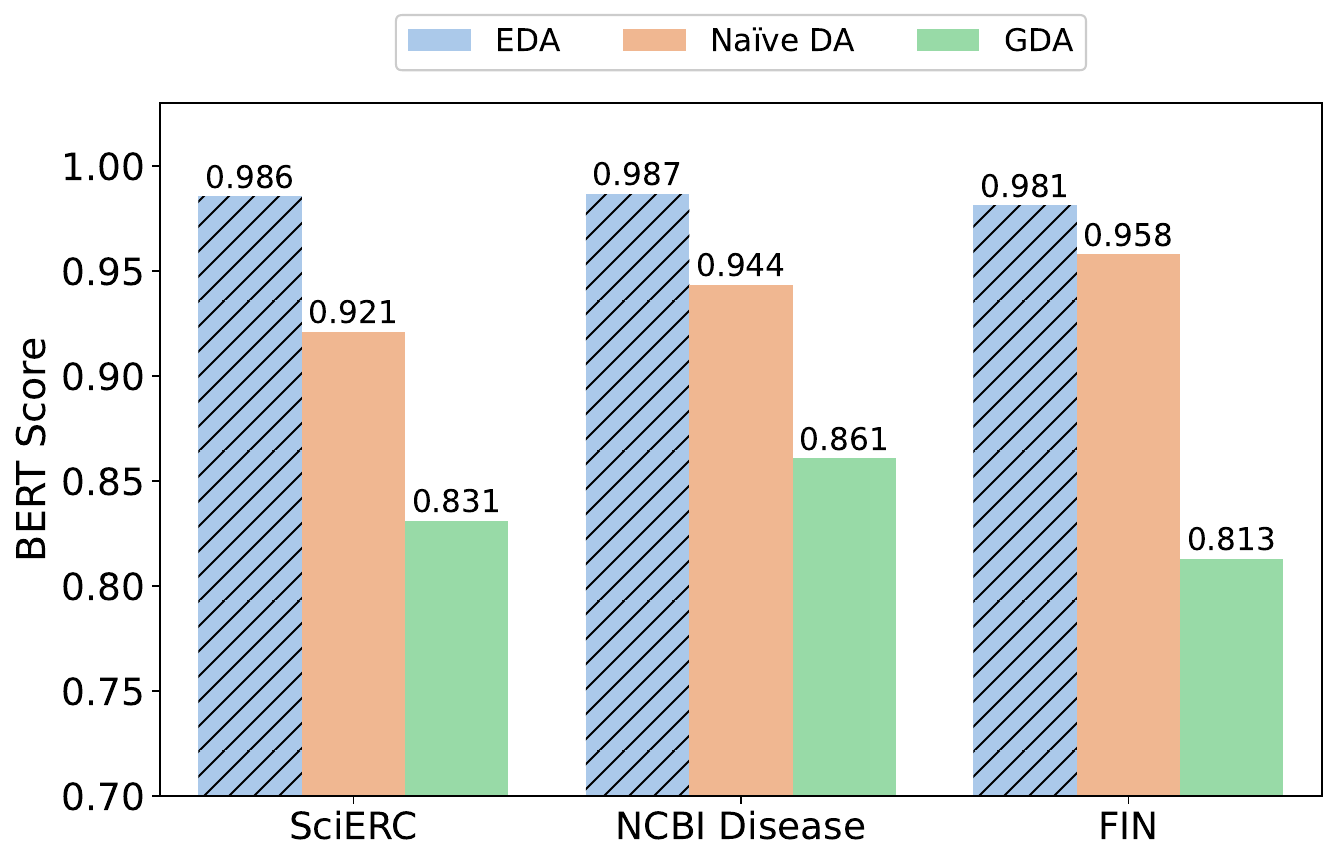}
        \caption{BERTScore in augmentation using GPT-4}
        \label{fig:bert_scoring_gpt4}
    \end{subfigure}
    \caption{Comparison of sentence diversity by method in augmentation using GPT-4 model. Lower BLEU and BERTScore indicate higher diversity in generated sentences.}
    \label{fig:scoring_comparison_gpt4}
\end{figure}

\section{Data Augmentation Details}\label{sec:baseline}
Experiments were conducted by categorizing DA methods into rule-based DA and LLM-based DA. For rule-based DA, WordNet and EDA were employed, with the number of synonyms to randomly choose set at 10. LLM-based DA encompassed Naïve DA and GDA, both utilizing identical settings.

\section{Implementation Details} \label{sec:imple_detail}
For the evaluation of DA, utilizing a learning rate of 2e-5, a batch size of 32, a maximum sequence length of 128, and the Adam optimizer as hyperparameters. The implementation framework utilized is based on Huggingface PyTorch Transformers \cite{huggingface}. In terms of computational infrastructure, the experimental procedures were exclusively executed on Nvidia A6000 GPUs, complemented by AMD CPU cores.

\section{Additional Sentence Diversity Evaluation}
\label{sec:gpt4_bleu}
Figure \ref{fig:bleu_scoring_gpt4} exhibits a graph that contrasts the sentence structures generated via various DA methods, employing the BLEU-4 metric for comparison. Data augmentation with GDA using GPT-4 exhibits a marked improvement in data diversity as opposed to GPT-3.5. Figure \ref{fig:bert_scoring_gpt4} shows that GPT-4 demonstrates enhanced semantic diversity in data augmentation using GDA. Both implementations of GDA, using GPT-3.5 and GPT-4, showed improvements in semantic diversity within data augmentation compared to other augmentation methods.


\section{Guidance Data Augmentation Case}
\label{sec:DA_case}
Table \ref{tab:da_case} presents an example of data created through EDA, Naïve DA, and GDA methods, with the last two utilizing the GPT-3.5 model. Augmentations via EDA and Naïve DA methods reveal replacements limited to either entity-specific words or other words. In contrast, sentences generated through GDA exhibit diversification in both entity-related vocabulary and overall sentence structure, while maintaining the context of the seed data despite structural modifications.

\end{document}